# DCA for Bot Detection

Yousof Al-Hammadi, Uwe Aickelin and Julie Greensmith

*Abstract*— Ensuring the security of computers is a non-trivial task, with many techniques used by malicious users to compromise these systems. In recent years a new threat has emerged in the form of networks of hijacked zombie machines used to perform complex distributed attacks such as denial of service and to obtain sensitive data such as password information. These zombie machines are said to be infected with a 'bot' - a malicious piece of software which is installed on a host machine and is controlled by a remote attacker, termed the 'botmaster of a botnet'. In this work, we use the biologically inspired Dendritic Cell Algorithm (DCA) to detect the existence of a single bot on a compromised host machine. The DCA is an immune-inspired algorithm based on an abstract model of the behaviour of the dendritic cells of the human body. The basis of anomaly detection performed by the DCA is facilitated using the correlation of behavioural attributes such as keylogging and packet flooding behaviour. The results of the application of the DCA to the detection of a single bot show that the algorithm is a successful technique for the detection of such malicious software without responding to normally running programs.

## I. INTRODUCTION

Computer systems and networks come under frequent attack from a diverse set of malicious programs and activity. Computer viruses posed a large problem in the late 1980's and computer worms were problematic in the 1990s through to the early 21st Century. While the detection of such worms and viruses is improving a new threat has emerged in the form of *the botnet*. Botnets are decentralised, distributed networks of subverted machines, controlled by a central commander, affectionately termed the 'botmaster'. A single bot is a malicious piece of software which, when installed on an unsuspecting host, transforms host into a zombie machine. Bots can install themselves on host machines through several different mechanisms, with common methods including direct download from the internet, through malicious files received as emails or via the exploitation of bugs present in internet browsing software [15].

Bots typically exploit traditional networking protocols for the communication component of their 'command and control' structure. Such variants of bots IRC (Internet Relay Chat) bots, HTTP bots and more recently Peer-to-Peer bots. In this research we are primarily interested in the detection of IRC bots as they appear to be highly prevalent within the botnet community, and seemingly little research has been performed within this area of computer security. IRC is a chat based protocol consisting of various 'channels' to which a user of the IRC network can connect. Upon infection of a host, the bot connects to the IRC server and joins the specified channel waiting for the attacker's commands. The bot is programmed to respond to various commands generated by the attacker through a Command and Control (C&C) structure using the IRC protocol [13]. In addition to the flexibility offered by IRC in the management and control of bots, this protocol is ideal for such attackers as it provides a high degree of anonymity for the attacker/botmaster. In early implementations, bots were used to perform distributed denial of services attacks (DDoS) using a flood of TCP SYN, UDP or ICMP 'ping' packets in an attempt to overload the capacity of computing resources. More recently bots are developed complete with keylogging features for closely monitoring user behaviour including the interception of sensitive data such as passwords, monitoring mouse clicks and the taking of screenshots of 'secure' websites [11].

In this paper we focus on the detection of a single bot formulated as a host-based intrusion detection problem, and avoids the technical problems of administrating a highly infective network within an academic environment. To perform this research, we rely on principles of 'extrusion detection' [2] where we do not attempt to prevent the bot from gaining access to the system, but we detect it as it attempts to operate and subvert the infected host. In addition to the monitoring of potential keylogging activity through keystroke analysis, network information is also incorporated into the implemented method of detection.

The algorithm used for the detection of a single active bot is the immune-inspired Dendritic Cell Algorithm (DCA) [9]. This algorithm is a '2nd Generation' Artificial Immune System (AIS), and is based on an abstract model of the behaviour of dendritic cells (DCs) [16]. These cells are the natural intrusion detection agents of the human body, whom activate the immune system in response to the detection of damage to the host. As an algorithm, the DCA performs multi-sensor data fusion on a set of input signals, and this information is correlated with potential 'suspects' which we term 'antigen'. This results in a pairing between signal evidence and antigen suspects, leading to information which will state not only if an anomaly is detected, but in addition the culprit responsible for the anomaly. Given the success of this algorithm at detecting scanning activity in computer networks [8] the DCA is a good candidate solution for the detection of a single bot.

The aim of this paper is to apply the DCA to the detection of a single bot and to assess its performance on this novel problem area. For these experiments the basis of classification is facilitated through the correlation of malicious activities such as keylogging, SYN or UDP flooding attacks, anomalous file access and potential bot-related communications. Our results show that correlating the

Yousof Al-Hammadi, Uwe Aickelin and Julie Greensmith are with the Department of Computer Science, The University of Nottingham, Jubilee Campus, Nottingham NG8 1BB, UK (email: yxa@cs.nott.ac.uk, uxa@cs.nott.ac.uk,jqg@cs.nott.ac.uk).

different behaviours exhibited by a single bot can enhance the detection of malicious processes on the system to determine the presence of a bot infection and to identify the processes involved in the bot's actions. As part of this investigation we also introduce a more sophisticated analysis method for the DCA which may give an improved performance than the anomaly coefficient method described in Greensmith *et al.* [8].

This paper is structured as follows: Section II contains background information describing the DCA algorithm. Section III discusses the existing bots detection techniques. We present our methodology of bot detection in section IV. The conducted experiments are explained in section V. Our results and analysis are presented in section VI and we summarize and conclude in section VII.

## II. THE DENDRITIC CELL ALGORITHM

### A. Algorithm Overview

Artificial Immune Systems have been applied to problems in computer security since their initial development in the mid-1990's. A recent addition to the AIS family is the DCA, which unlike other AISs does not rely on the pattern matching of strings (termed antigen), but instead uses principles from the danger theory to perform 'context aware' intrusion detection. The danger theory is an alternative view of presenting the activation of the immune system. The danger theory states a response is generated by the immune upon the receipt of molecular signals which indicate that the body is in distress. DCs are sensitive to changes in the concentration of danger signals. In this work we have produced an abstract view of the essential features of DC biology, which are presented in this paper. For further information on the natural functions of DCs please refer to Lutz and Schuler [14]. A detailed description and formal analysis of the DCA is given in Greensmith *et al.* [9]

In addition to the processing of signals, DCs consume any proteins within their locality and storing these antigen proteins for future use. DCs combine the evidence of damage with the collected suspect antigen to provide information about how 'dangerous' a particular protein is to the host body. In addition to danger signals, two other sources of signal are influential regarding the behaviour of DCs - namely PAMPs (pathogen associated molecular patterns) and safe signals which are the molecules released as a result of normal cell death.

In nature DCs exist in three states: immature, semi-mature and mature. The initial state of a DC is immature, where it performs its function of processing the three categories of input signal (PAMP, danger and safe) and in response produces three output signals. Two of the output signals are used to represent the state of the cell, as the immature DC can change state irreversibly to either semi-mature or mature. During its lifespan collecting signals, if the DC has collected majority safe signals it will change state to a semi-mature state and all antigen collected by the cell is presented in a 'safe' context. Conversely, cells exposed to PAMP and danger signals transforms into a mature state, with all collected antigen presented in a dangerous context.

TABLE I
SIGNAL WEIGHT VALUES

|      | csm | semi | mat |
|------|-----|------|-----|
| PAMP | 4   | 0    | 8   |
| DS   | 2   | 0    | 4   |
| SS   | 3   | 1    | -6  |

To initiate maturity, a DC must have experienced signals, and in response to this express output signals. As the level of input signal experienced increases, the probability of the DC exceeding its lifespan also increases. The level of signal input is mapped as a *costimulatory output signal* (CSM). Once CSM reaches a 'migration' threshold value, the cell ceases signal and antigen collection and is removed from the population for analysis. Upon removal from the population the cell is replaced by a new cell, to keep the population level static. Each DC is assigned a different migration threshold. This results in a population of cells whom sample for different durations and experience different input signal combinations. The transformation from input to output signal per cell is performed using a simple weighted sum (Equation 1) described in detail in [8] with the corresponding weights given in Table I, with values derived from preliminary experimentation. Pseudocode for the functioning of a single cell is presented in Algorithm 1.

---

**input** : Sorted antigen and signals (PAMP,DS,SS)
**output**: Antigen and their context (0/1)

Initilize DC;
**while** *CSM output signal < migration threshold* **do**
  *get antigen*;
  *store antigen*;
  *get signals*;
  *calculate interim output signals*;
  *update cumulative output signals*;
**end**
cell location update to lymph node;
**if** *semi-mature output > mature output* **then**
  *cell context is assigned as 0*;
**else**
  *cell context is assigned as 1*;
**end**
kill cell;
replace cell in population;

**Algorithm 1**: DCA algorithm

---

$$O_j = \sum_{i=0} (W_{ijk} * S_i) \qquad \forall j, k \qquad (1)$$

where:
- $W$ is the signal weight of the category $i$
- $i$ is the input signal category ($i_0 = PAMP$, $i_1 = DS$, and $i_2 = SS$)

- $k$ is the weight set index as shown in Table II ($k = 1 \ldots 5$)
- $j$ is the output signal value where:
  - $j = 0$ is a costimulatory signal (csm)
  - $j = 1$ is a semi-mature DC output signal (semi)
  - $j = 2$ is a mature DC output signal (mat)

*B. Signals and Antigen*

In nature the three different categories of signal are derived from different sources and have different effects on DCs. To use such signals within a computational context, abstract semantic mappings between potential sources of input data and the signal categories are derived. The signal mappings are as follows, with the details of the signals used for the detection of bots given in section IV:

- PAMPs: A signature of abnormal behaviour. An increase in this signal is associated with a high confidence of abnormality.
- Danger Signal: A measure of an attribute which increases in value to indicate an abnormality. Low values of this signal may not be anomalous, giving a high value a moderate confidence of indicating abnormality.
- Safe Signal: A measure which increases value in conjunction with observed normal behaviour. This is a confident indicator of normal, predictable or steady-state system behaviour. This signal is used to counteract the effects of PAMPs and danger signals.

In previous experiments with the DCA, the system calls invoked by running processes are used as antigen [7]. This implies that behavioural changes observed within the signals are potentially caused by the invocation of running programs. For the purpose of bot detection, antigen are derived from API function calls, which are similar to system calls. The resultant data is a stream of potential antigen suspects, which are correlated with signals through the processing mechanisms of the DC population. One constraint on antigen is that more than one of any *antigen type* must be used to be able to perform the anomaly analysis with the DCA. This will allow for the detection of which type of function call is responsible for the changes in the observed input signals. More details are given in Section IV-D regarding the mapping of antigen.

*C. Analysis*

Once all antigen and signals are processed by the cell population, an analysis stage is performed. This stage involved calculating an anomaly coefficient per antigen type - termed the *mature context antigen value*, MCAV. The derivation of the MCAV per antigen type in the range of *zero* to *one* is shown in Equation 2. The closer this value is to one, the more likely the antigen type is to be anomalous.

$$MCAV_x = \frac{Z_x}{Y_x} \quad (2)$$

where $MCAV_x$ is the MCAV coefficient for antigen type $x$, $Z_x$ is the number of mature context antigen presentations for antigen type $x$ and $Y_x$ is the total number of antigen presented for antigen type $x$.

In previous work [9], it has been shown that the MCAV for processes with low numbers of antigen per antigen type can be higher than desired. This can lead to the generation of false positives. In this paper we address this problem by producing an anomaly coefficient which is an improvement on the MCAV, by incorporating the number of antigen used to calculate the MCAV. This improvement is termed the *MCAV Antigen Coefficient* or MAC. The MAC value is the MCAV of each antigen type multiplied by the number of output antigen per process and divided by the total number of output antigen for all processes. This calculation is shown in Equation 3. The MAC value also ranges between *zero* and *one*. As with the MCAV, the nearer the MAC value to *one*, the more anomalous the process.

$$MAC_x = \frac{MCAV_x * Antigen_x}{\sum_{i=1}^{n} Antigen_i} \quad (3)$$

where $MCAV_x$ is the MCAV value for process $x$ and $Antigen_x$ is the number of antigen processed by process $x$.

III. RELATED WORK: BOTS AND THEIR DETECTION

*A. Bots*

The aim of a bot is to subvert a host machine for use by the central controller. In order to achieve this aim, each bot is armed with various methods to facilitate their malicious activities. To communicate with the bot commander, the IRC bot must connect to an IRC channel, where the bot is termed to have 'membership'. Once a bot is installed and becomes active on a channel, it awaits commands from its controller. Bots frequently rely on the ability to perform keylogging. This is the means of intercepting and recording user activities such as keystroke typing. Keylogging represents a serious threat to the privacy and security of our systems as the keylogger can collect sensitive information from the user such as personal information, passwords and credit card numbers. The acquisition of this information can constitute identity theft and fraud.

Another malicious activity is performing a denial of service attack by sending numerous network packets to the remote host across the network, including both SYN and UDP packet flooding. SYN attacks are invoked when the controlling botmaster issues a SYN attack command to the bot and exploits the '3-way handshake' of a TCP connection stream. The bot on the infected host sends a series of SYN packets to the target host, using modified packet headers to disguise the IP address of the infected host. The target host replies with a SYN ACK packet. The target host then waits to receive the corresponding ACK from the bot-infected machine. However, the bot never responds back because the SYN ACK is sent to a random IP address as the bot had spoofed the IP address of the outbound packets. As a result, multiple connection requests accumulate at the target host, resulting in the victim memory buffers become full so that it cannot accept further legitimate connection requests causing a denial of service, as the target is rendered unusable. In a similar manner, UPD floods are also used to slow down a target system to the point where further connections cannot be handled, by sending large number of UDP packets to a specified port on a remote host.

In addition to the aforementioned attacks, the infection of a bot on a host machine seriously compromises the confidentiality of the data contained within the infected host. In effect, once a bot has infected a machine, it becomes under the control of the botmaster and can be subverted for whatever purpose the controller requires.

*B. Detection Techniques*

The majority of existing techniques for botnet detection are signature-based approaches, in a similar manner to many intrusion detection systems. Such techniques frequently use the analysis of network traffic [4][5][6]. Although this approach is a useful mechanism for bot detection, it is impossible if the network packet data is encrypted. Freiling *et al.* [5][10] collect bot binaries by using a non-productive resource, such as a honeypot, to analyse bot traffic and collect useful information to shut down the remote control network by emulating bot activities. There are numerous problems with this approach. First, the non-productive resource needs to receive activities directed against it in order to analyse bot behaviour. In addition, emulating bots' actions to join botnet community can be discovered if the botnet size is relatively small. To avoid these problems, our work focuses on monitoring API function calls.

The technique used by Cooke et al. [4] performs bot detection through the analysis of two types of bot-based communications, namely bot-to-bot and between bots and their controllers. In the approach taken by Cooke *et al.*, bot payloads are analysed using pattern matching of known bot commands and in addition examines a system for evidence of non-human characteristics. They conclude that bots can run on non-standard ports and that the analysis of encoded packets is very costly on high throughput networks. They determined that there are no simple characteristics of the bot communication channels that can be used for detection, which makes the detection of bots an interesting and difficult problem. They also discuss the approach of detecting bots by their means of distribution or attack behaviour by correlating data from different sources. While it is suggested in their work that correlation would be beneficial for the detection of a single bot, they did not provide information regarding how this correlation should be performed. To alleviate this problem, the DCA is applied to bot detection, with their suggestions further promoting the use of such correlation algorithms for this type of detection problem.

A alternative method for detecting bots introduced by Goebel and Holz [6] through monitoring IRC traffic for suspicious IRC nicknames, IRC servers and non-standard server ports. Such potentially suspicious packets are assessed using regular expressions to classify suspicious nicknames, resulting in a scoring function per packet. However, their approach can be overcome in numerous ways such as using hitlists which contain normal names or again, through the encryption of such vital information.

Anomaly detection is also used to detect the presence of a bot [3], where deviations from a defined 'normal' are classed as an anomaly. In anomaly detection, behavioural attributes are often profiled to perform the assessment of potentially anomalous data. An approach for detecting bots using behavioural analysis is presented by Racine [15]. This method is based on the discovery of inactive clients and their subsequent assignment to a network connection. Any active clients are then classified according to the IRC channel membership. This approach is successful in detecting idle IRC activity, but suffers from high false positive rates when applied to a scenario consisting of both active users and active bots. As with similar techniques, searching for such IRC patterns can be costly and difficult especially if the packets are encrypted. However, we believe the use of behavioural monitoring is a promising method for the detection of bots, especially if such attributes can be used to correlate the behavioural changes with evidence of active IRC bots. This provides us with a strong motivation for using an algorithm designed for the purpose of correlating behaviour with activity, as performed by the DCA.

In summary the majority of techniques for the detection of a single bot are based on developing signatures and through the use of network packet header information. These techniques are limited - if packet streams are encrypted then these measures can be circumvented with relative ease. Current behaviour-based approaches are also limited, generating high rates of false positives, which have the potential to slow down a system. We believe that using the DCA to perform this detection task will be successful as it contains elements of signature based detection through the use of the PAMP signals, which is combined with the anomaly-based approach represented by the danger signals. In addition the DCA can be used to correlate relevant behavioural attributes with programs potentially involved with a bot infection.

## IV. METHODOLOGY

*A. Bot Scenarios*

For the purpose of experimentation two different types of bot are used, namely spybot and sdbot [1]. The spybot is a suitable candidate bot as it uses various malicious functionalities such as keylogging and SYN attacks. The sdbot is also used as it contains the additional functionality of a UDP attack. As a communication vessel, IceChat [12], an IRC client, is used for normal conversation and to send files to a remote host. To provide suitable data for the DCA a 'hooking' program is implemented to capture the required behavioural attributes and to intercept and capture function calls. To emulate real-world bot infections, three different scenarios are constructed including inactive (E1), attack (E2.1-2.3) and normal (E3) scenarios. The attack scenario consists of three sessions: a keylogging attack session, a flooding session and a combination session comprising both

keylogging and packet flooding. The derived sessions include the following:

- *Inactive bot (E1):* This session involves having *inactive bots* running on the monitored host in addition to normal applications such as an IRC client, Wordpad, Notepad and CMD processes. Spybot is used for this session. The bot runs on the monitored victim host and connects to an IRC server where it joins a specified channel to await commands from its controller, though no attacking actions are performed by this idle bot. This results in minimal data, with the majority of transactions involving simple PING messages between the bot, the IRC server and the IceChat IRC client.
- *Keylogging Attack (E2.1):* The sdbot is capable of intercepting keystrokes, upon receipt of the relevant command from the botmaster. Bots use various methods to perform keylogging - both techniques involves the bot intercepting API (Application Programming Interface) function calls. In this scenario, two methods of keylogging are used including the "GetKeyboardState" and "GetAsyncKeyState" function calls. However, detection cannot be performed by examining these two function calls alone, as normal legitimate programs often rely on such function calls. For example, MS Notepad utilises GetKeyboardState as part of its normal functioning. The DCA will be employed to discriminate between malicious and legitimate keystroke function calls.
- *Flooding Attack (E2.2):* This involves performing packet flooding using the spybot for a SYN flood attack and the sdbot for a UDP attack. These flooding methods are designed to emulate the behaviour of a machine partaking in a distributed denial of service attack. As part of the process of packet flooding the bots rely heavily on socket usage, as part of the packet sending mechanism. Therefore to detect these attacks, socket uses monitors are employed, with the exact nature of this data given in the forthcoming section. It is important to note that during the flooding attack no 'normal' legitimate applications are running.
- *Combined Attack (E2.3):* In this session, both keylogging and SYN flooding are invoked by the bot. As with session E1, spybot is used to perform this attack. Note that the two activities can occur simultaneously in this scenario.
- *Normal Scenario (E3):* The normal scenario involves transferring a file of 10 KB from one host to another through IRC client. Other applications such as Wordpad, Notepad, cmd and the hook program are running on the victim host. Note that no bots are used in this scenario as this is the normal/uninfected session used to assess the occurrence of any potential false positive errors made by the DCA.

### B. Data Collection

It is assumed that either bot is already installed on the victim host, through an accidental 'trojan horse' style infection mechanism. Therefore we are not attempting to prevent the initial bot infection but to limit its activities whilst on a host machine. The bot runs as a process whenever the user reboots the system and attempts to connect to the IRC server through IRC standard ports (in the range of 6667-7000). The bot then joins the IRC channel and waits for the botmaster to login and issue commands.

For the purpose of use by the DCA an interception program is implemented and run on the victim machine to collect the required signals and antigen data. These collected data are then processed, normalised and streamed to the DCA. In terms of the function calls intercepted, three specific types of function calls are used as signal and antigen input to the DCA. These function calls are as follows:

- Communication functions: socket, send, sendto, recv and recvfrom.
- File access functions: CreateFile, OpenFile, ReadFile and WriteFile.
- Keyboard state functions: GetAsyncKeyState, GetKeyboardState, GetKeyNameText and keybd_event.

Invoking these function calls within specified time-window can represents a security threat to the system, but also may form part of legitimate usage. Therefore, an intelligent correlation method such as the DCA is required to determine if the invocations of such function calls are indeed anomalous.

### C. Signals

Signals are mapped as a reflection of the state of the victim host. A major function of the DC is the ability to combine signals of different categories to influence the behaviour of the artificial cells. Three signal categories are used to define the state of the system namely PAMPs, DSs (danger signals) and SSs (safe signals), with one data source mapped per signal category. These signals are collected using a function call interception program. Raw data from the monitored host are transformed into log files which are then analysed by the DCA, following a signal normalisation process. The resultant normalised signals are in the range of 0 - 100 for the PAMP and DS with the SS having a reduced range, as suggested in Greensmith *et al* [8]. This reduction ensures that the mean values of each signal category are approximately equal, with preliminary experiments performed to verify this.

In terms of the signal category semantics, the PAMP signal is a signature based signal. This signal is derived from the rate of change of invocation of selected API function calls used for keylogging activity. Such function calls include GetAsyncKeyState, GetKeyboardState, GetKeyNameText and keybd_event when invoked by the running processes. To use this data stream as signal input, the rate values are normalised. For this process $n_p$ is defined as the maximum number of function calls generated by pressing a key within one second. Through preliminary experimentation it is shown that by pressing any key on the keyboard for a one second it generates $n_p$ number of calls. Subsequently $n_p$ is set to be the maximum number of calls that can be generated per second. The resultant normalised the PAMP signal based on this value by applying linear scale between 0 and 100.

The danger signal is derived from the time difference between receiving and sending data through the network. As bots respond directly to botmaster commands, a small time difference between sending and receiving data is observed. In contrast, normal chat will have a higher value of time difference between sending and receiving activity. As with the PAMP signal, the normalisation of the danger signal involves calculating a maximum value. For this purpose $n_d$ is the maximum time difference between sending a request and receiving a feedback. If the time difference exceeds $n_d$, the response time is normal. Otherwise, the response time falls within the abnormality range and is scaled between 0 and 100 through the use of this maximum value. We set up a critical range (0 to $n_d$) that represents a abnormal response time. The zero value is mapped to 100 (danger) and $n_d$ is mapped to zero (not danger). If the response time falls within the critical value, it means that the response is fast and consider to be dangerous.

Finally, the safe signal is derived from the time difference between two outgoing consecutive communication functions such as [(send,send),(sendto,sendto),(socket,socket)]. This is needed as the bot sends information to the botmaster or issues SYN or UDP attacks. In normal situation, we expect to have a large time difference between two consecutive functions. In addition, we expect to have a short period of this action in comparison to SYN attack or UDP attack. Therefore $n_{s1}$ and $n_{s2}$ are the maximum time differences between calling two consecutive communication functions. If the time difference is less than $n_{s1}$, the time is classified as abnormal. If the time difference falls between $n_{s1}$ and $n_{s2}$, the time difference is classified as uncertain time. If the time difference is more than $n_{s2}$, the time difference is classified as safe time. These timings are scaled between 0 and 10. Using statistical analysis, we notice that the mean value for bot to respond to the command is around 3.226 seconds. Therefore, we set up a critical range for the safe signal. We divide our critical range into three sub-ranges. The first range is from *zero* to $n_{s1}$ where $n_{s1} = 5$. Any value that falls within this range is considered as an unsafe signal. The second range is where there is uncertainty of response. The uncertainty range is between $n_{s1}$ and $n_{s2} = 20$. The third range is that the time difference is above $n_{s2}$ and is considered as the confidence range. In this range, we are sure that the time difference between two consecutive function calls is generated as a normal response.

### D. Antigen

The collected signals are a reflection of the status of the monitored system. Therefore, antigen are potential culprits responsible for any observed changes in the status of the system. In order to detect the bot, correlation of antigen signals is required to define which processes are active when the signal values are modified. Any process executed one of the selected API function calls explained in section IV, the process id which causes the calls is stored as an antigen in the antigen log file. The more active the process, the more antigen it generates. Each intercepted function call is stored and is assigned the value of the process ID to which the function call belongs and the time at which it is invoked. Both signal and antigen logs are combined and sorted based on time. The combined file form a dataset which is passed to the DCA through a data processing client. The combined log files are parsed and the logged information is sent to the DCA for processing and analysis.

TABLE II
WEIGHT SENSITIVITY ANALYSIS

|  | signal | WS1 | WS2 | WS3 | WS4 | WS5 |
|---|---|---|---|---|---|---|
| csm | PAMP | 2 | 4 | 4 | 2 | 8 |
|  | DS | 1 | 2 | 2 | 1 | 4 |
|  | SS | 2 | 6 | 3 | 1.5 | 0.6 |
| semi | PAMP | 0 | 0 | 0 | 0 | 0 |
|  | DS | 0 | 0 | 0 | 0 | 0 |
|  | SS | 1 | 1 | 1 | 1 | 1 |
| mat | PAMP | 2 | 8 | 8 | 8 | 16 |
|  | DS | 1 | 4 | 4 | 4 | 8 |
|  | SS | -3 | -12 | -6 | -6 | -1.2 |

## V. EXPERIMENTS

The aim of these experiments is to use the DCA to perform detection of the bot. Various experiments are performed to verify this aim. Each experiment is repeated ten times which is sufficient, as the results from the repeated experiments produce a small variation on standard deviation by using Chebyshev's Inequality. After collecting and processing the data, one dataset is selected randomly from each repeated experiments. The dataset is passed to the DCA using a data collection client. Each scenario (E1 - E3) is used for the purpose of experimentation with three hypotheses tested:

1) *Null Hypothesis One:* Data collected per dataset are normally distributed. The Shaprio Wilk test is used for this assessment.
2) *Null Hypothesis Two:* The MCAV/MAC values for the normal processes are not statistically different from those produced by the bot process. This is verified through the performance of a two-sided Mann-Whitney test, where $p = 0.95$.
3) *Null Hypothesis Three:* Variation of the signal weights as described in Table II produces no observable difference in the resultant MCAV/MAC values and the detection accuracy. Wilcoxon signed rank tests (two-sided) are used to verify this hypothesis.

### A. System Setup

In all experiments, the parameters used are identical to those implemented in Greensmith and Aickelin [8], with the exception of the weights, as per Null Hypothesis Three. All experiments are performed in a small virtual IRC network on a VMWare workstation. The VMWare workstation runs under a Windows XP P4 SP2 with 2.4 GHz processor. The virtual IRC network consists of two machines, one IRC server and one infected host machine. Two machines are sufficient to perform these experiments as one host is required to be infected (i.e. the monitored machine) and the other to be an IRC server to issue commands to the bot in question.

TABLE III
THE RESULTS OF THE MCAV/MAC VALUES GENERATED FROM DCA BASED ON SIGNAL WEIGHTS OF TABLE I. THE VALUES THAT HAVE ASTERISKS ARE NOT SIGNIFICANT

| Experiment | Process | Output Antigen | mean | | Mann-Whitney P-Value | |
|---|---|---|---|---|---|---|
| | | | MCAV | MAC | MCAV | MAC |
| E1 | bot | 35 | 0.0978 | 0.0578 | | |
| | IRC | 24 | 0.0625 | 0.0255 | 0.1602* | 0.0202 |
| E2.1.a | bot | 1329.7 | 0.4736 | 0.4542 | | |
| | IRC | 59 | 0.2881 | 0.0122 | 0.0002 | 0.0002 |
| E2.1.b | bot | 1296.2 | 0.5441 | 0.3098 | | |
| | IRC cmd | 464.9 | 0.5284 | 0.1077 | 0.0089 | 0.0002 |
| | Notepad | 8.9 | 0.7889 | 0.0031 | 0.0002 | 0.0002 |
| | Wordpad | 239.4 | 0.6916 | 0.0726 | 0.0002 | 0.0002 |
| | | 268.8 | 0.8286 | 0.0977 | 0.0002 | 0.0002 |
| E2.2.a | bot | 19206.3 | 0.6047 | 0.6038 | | |
| | IRC | 18 | 0.3441 | 0.0003 | 0.0002 | 0.0000 |
| | cmd | 9.8 | 0.2889 | 0.0002 | 0.0003 | 0.0000 |
| E2.2.b | bot | 5790.5 | 0.4360 | 0.4346 | | |
| | IRC | 19 | 0.2772 | 0.0009 | 0.0002 | 0.0000 |
| E2.3.a | bot | 41456 | 0.8218 | 0.8214 | | |
| | IRC | 20.5 | 0.5480 | 0.0003 | 0.0002 | 0.0000 |
| E2.3.b | bot | 22446 | 0.9598 | 0.9461 | | |
| | IRC cmd | 59.1 | 0.7802 | 0.0021 | 0.0000 | 0.0002 |
| | Notepad | 9.7 | 0.6300 | 0.0003 | 0.0002 | 0.0002 |
| | Wordpad | 23.1 | 1.0000 | 0.0010 | 0.0001 | 0.0002 |
| | | 233.6 | 0.8801 | 0.0090 | 0.0002 | 0.0002 |
| E3 | IRC | 135.5 | 0.1136 | 0.1136 | N/A | N/A |

## VI. RESULTS AND ANALYSIS

The results from the experiments are shown in Table III,IV and V. The mean MCAV and the mean MAC values for each process are presented, derived across the ten runs performed per scenario.

### A. Null Hypothesis One

Upon the application of the Shapiro-Wilk test to each of the datasets, it was discovered that the resultant p-values imply that the distribution of scenarios E1, E2.1 and E3 are normal, with E2.2 and E2.3 not normal. Given these two rejections of the null, Null Hypothesis One is rejected. As a result of this, further tests with these data use non-parametric statistical tests such as the Mann-Whitney test, also using 95% confidence.

### B. Null Hypothesis Two

For all scenarios E1-E3, a comparison is performed using the results generated for the bot versus all other normal processes within a particular session. The results of this comparison are presented in Table III. In this table, the computed p-values using an unpaired Mann-Whitney test are presented, with those results deemed not statistically significant marked with an asterisks. In experiment E1, significant differences do not exist between the resultant MCAV values for the inactive bot and the normal IRC process, and so for this particular scenario the null cannot be rejected. This can be attributed to the fact that the total number of antigen produced by both processes is too small in number to give an accurate description of the state of the monitored host. This is supported by the fact that the MAC values differ significantly for this experiment. This implies that the MAC is a useful addition to the analysis of the data output from the DCA as it allowed for the incorporation of the antigen data, which can influence the interpretation of the results.

Significant differences are shown by the low p-values presented in Table III for experiments E2.1.a and E2.1.b for both the MAC and MCAV coefficient values, where the sample size is equal to ten. The differences are further pronounced in the generation of the MAC values, further supporting its use with the DCA. We can conclude therefore, that the DCA can be used in the discrimination between normal and bot-directed processes and that the DCA is successful in detecting keylogging activities. This trend is also evident for scenarios E2.2.a/b and E2.3.a/b, where the bot process MAC and MCAV are consistently higher than those of the normal processes, IRC and notepad inclusive. This information is also displayed in Figures 1 and 2. This implies that in addition to the detection of the bot itself the DCA can detect the performance of outbound scanning activity - further supporting the experiments performed in Greenmsith and Aickelin [8]. Therefore the null hypothesis can be rejected as in the majority of cases the DCA successfully discriminates between normal and bot processes, with the exception of E1.

### C. Null Hypothesis Three

Tables IV and V include the results of the sensitivity analysis on the weight values for the bot process. The figures presented are mean values taken across the ten runs per session (E1-E2). An arbitrary threshold is applied at 0.5: values above deem the process anomalous, and below as normal. From these data, it is shown that changing the weights used in the signal processing equation has some influence on the performance of the system. For example, in the case of session E2.1.a, weight set (WS) WS1 produces a MAC value of 0.09 for the bot yet produces a value of 0.72 for WS5. This increase is likely to reduce the rate of false negatives. To further explore these effects, the resultant data are plotted as boxplots as the data are not normally distributed. To assess the performance of the DCA as an anomaly detector the results for the anomalous bot and the normal IRC client are shown for the purpose of comparison. For these boxplots, the central line represents the median value, with the drawn boxes representing the interquartile ranges.

In Figure 3 the median MCAV values are presented, derived per session across the ten runs performed for each WS (n=50). For the bot process, (Figure 3(a)) the MCAV is low for session E1, in-line with previous results. For E1, variation in the weights does not influence the detection results, as this process has low activity and therefore does not generate any great variation in the signals. Therefore, without input variation, the output does not vary in response to changing the manner in which the input is processed. This is also evident in Figure 4(a) when using the MAC value.

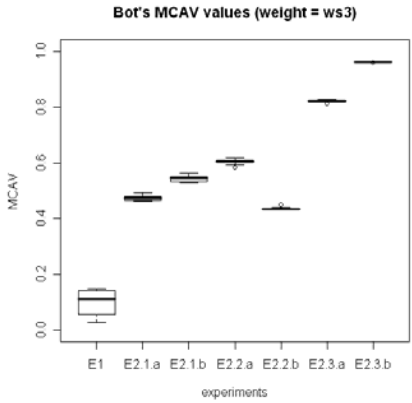
(a) Bot's MCAV value using weight set (WS3)

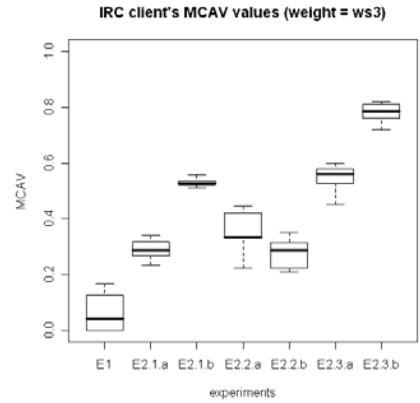
(b) IRC client's MCAV value using weight set (WS3)

Fig. 1. The MCAV values generated by DCA based on the weights described on Table I

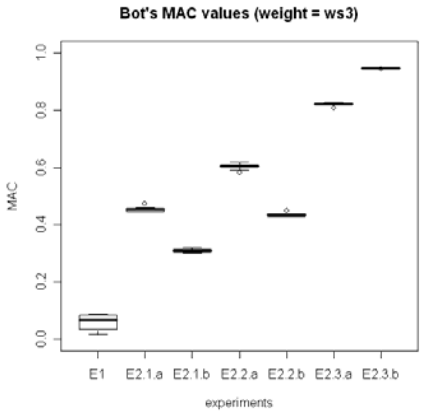
(a) Bot's MAC value using weight set (WS3)

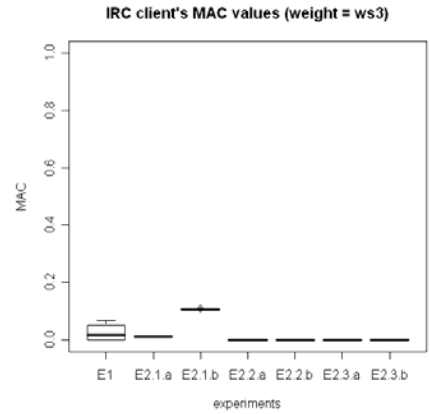
(b) IRC client's MAC value using weight set (WS3)

Fig. 2. The MAC values generated by DCA based on the weights described on Table I

For all other sessions, much greater variation is observed upon weight modification, as shown by the large interquartile ranges produced for both MCAV and MAC values of the bot processes. While the similar trends are shown across the sessions in the MCAV of the IRC client, differences are evident for the MAC value. In Figure 4(b) it is evident that all sessions have low MACs for this process across all weight sets. Therefore as the weights are modified, there is a greater influence on the anomalous processes than on the normal processes. Should the arbitrary threshold applied to the MAC values be set at 0.2 as opposed to 0.5, then the performance of the DCA on botnet detection is good, producing low rates of false positives and high rates of true positives.

To further explore this effect, an alternative plotting is presented in Figures 5 and 6. Here, each bar represents the results for each WS, derived from the ten runs per session (E1-E2 inclusive) totaling 70 runs per bar. As with Figures 3 and 4 some influence is shown through weight modification.

WS1 produced low MCAV and MAC values for both processes. This indicates that these weights, used previously with the DCA, are less suitable for this particular application. WS2 shows little improvement when compared to WS1, producing the lowest MCAV and MAC for the bot process. WS3 shows an improved performance, producing much higher MAC values for the bot process and very low values for the IRC client. WS4 produced even higher values for the bot process, whilst keeping the values low for the normal IRC client process. WS5 also produced high MCAV and MAC values for the bot, but the interquartile range of the normal process increased. This suggests that as the ratio of PAMP to safe signal weight for producing the mature output signal must be sufficiently small to avoid false positives, and sufficiently large to avoid potential false negatives.

Finally, to verify these findings statistically, each set of results per session per weight are compared exhaustively using the non-parametric Wilcoxon signed rank test. For each test performed the resultant p-value is less than 0.001. This allows us to conclude that modification of the weights has a significant effect on the output of the DCA when applied to this detection problem, and leads to the rejection of Null

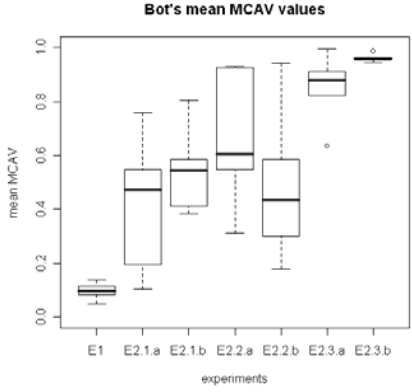
(a) Bot's mean MCAV value

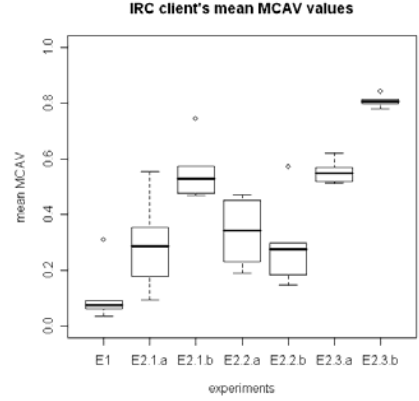
(b) IRC client's mean MCAV value

Fig. 3. The mean MCAV/MAC values generated by DCA using different signal weight values (WS1-WS5)

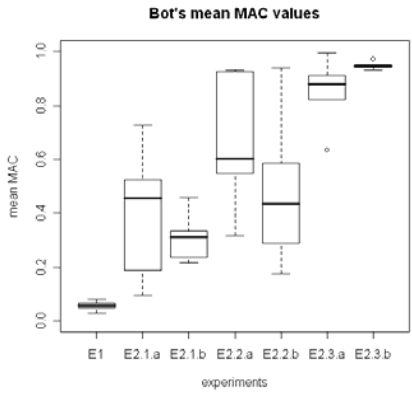
(a) Bot's mean MAC value

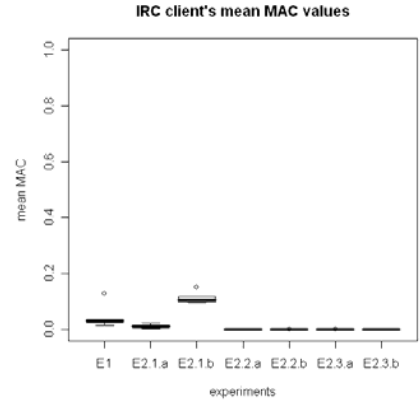
(b) IRC client's mean MAC value

Fig. 4. The mean MCAV/MAC values generated by DCA using different signal weight values (WS1-WS5)

Hypothesis Three.

TABLE IV
WEIGHT SENSITIVITY ANALYSIS FOR THE BOT'S MCAV VALUES

| Expriment | WS1 | WS2 | WS3 | WS4 | WS5 |
|---|---|---|---|---|---|
| E1 | 0.0484 | 0.0834 | 0.0978 | 0.1140 | 0.1377 |
| E2.1.a | 0.1030 | 0.1964 | 0.4736 | 0.5477 | 0.7595 |
| E2.1.b | 0.3823 | 0.4123 | 0.5440 | 0.5861 | 0.8032 |
| E2.2.a | 0.5488 | 0.3119 | 0.6047 | 0.9269 | 0.9319 |
| E2.2.b | 0.2995 | 0.1770 | 0.4360 | 0.5863 | 0.9427 |
| E2.3.a | 0.8802 | 0.6345 | 0.8218 | 0.9112 | 0.9955 |
| E2.3.b | 0.9553 | 0.9443 | 0.9598 | 0.9641 | 0.9873 |

TABLE V
WEIGHT SENSITIVITY ANALYSIS FOR THE BOT'S MAC VALUES

| Expriment | WS1 | WS2 | WS3 | WS4 | WS5 |
|---|---|---|---|---|---|
| E1 | 0.0288 | 0.0495 | 0.0578 | 0.0671 | 0.0810 |
| E2.1.a | 0.0947 | 0.1882 | 0.4543 | 0.5246 | 0.7274 |
| E2.1.b | 0.2168 | 0.2355 | 0.3098 | 0.3340 | 0.4572 |
| E2.2.a | 0.5479 | 0.3155 | 0.6038 | 0.9255 | 0.9305 |
| E2.2.b | 0.2886 | 0.1764 | 0.4345 | 0.5845 | 0.9395 |
| E2.3.a | 0.8798 | 0.6342 | 0.8214 | 0.9108 | 0.9949 |
| E2.3.b | 0.9418 | 0.9306 | 0.9461 | 0.9507 | 0.9726 |

## VII. CONCLUSION AND FUTURE WORK

In this paper, we have applied the DCA to the detection of a single bot with three null hypotheses explored. It is shown that the DCA is capable of discriminating between bot and normal processes on a host machine. Additionally, the incorporation of the MAC value has a positive effect on the results, significantly reducing false positives. Finally, the modification of the weights used in the signal processing component has a significant effect on the results of the algorithm. It is concluded that appropriate weights for this application include high values for the safe signal weight which appears to be useful in the reduction of potential false positives without generating false negative errors. We now intend to apply the DCA to the detection of "peer-to-peer" bots, which pose an interesting problem as the use of peer-to-peer networks increases. In addition we aim to use the results of these experiments to further our understanding of the DCA, to ultimately enhance the performance of this immune-inspired detection algorithm.

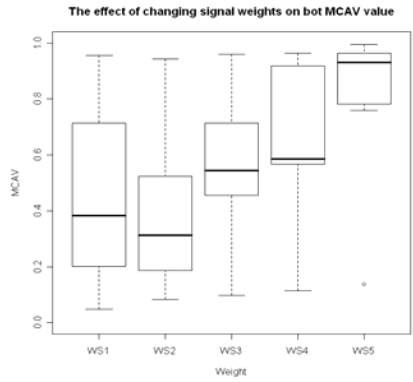
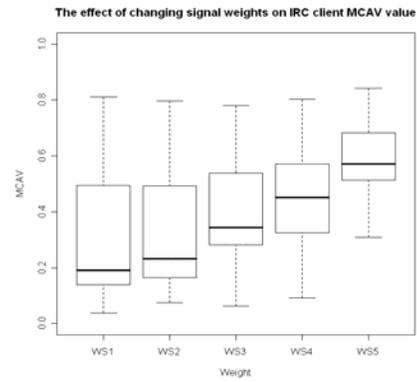

(a) Bot's mean MCAV value

(b) IRC client's MCAV value

Fig. 5. Affect of changing signal weights of the MCAV values on DCA detection performance

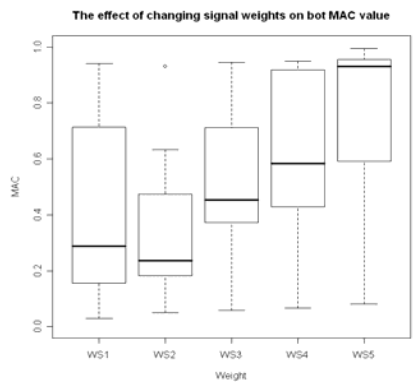
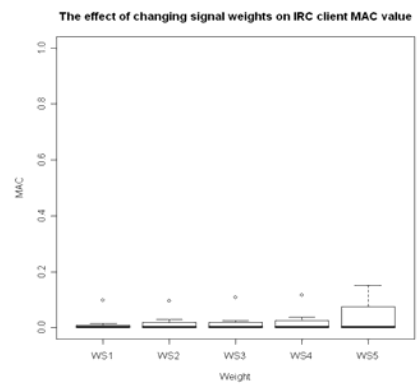

(a) Bot's MAC value

(b) IRC client's MAC value

Fig. 6. Affect of changing signal weights of the MAC values on DCA detection performance

ACKNOWLEDGMENT

The authors would like to thank Etisalat College of Engineering and Emirates Telecommunication Corporation (ETISALAT), United Arab Emirates, for providing financial support for this work.

REFERENCES

[1] P. Barford, and V. Yegneswaran, "An Inside Look at Botnets", Special Workshop on Malware Detection Advances in Information Security, Springer Verlag, 2006.
[2] R. Bejtlich, "Extrusion Detection: Security Monitoring for Internal Intrusions", Addison-Wesley Professional, 2005.
[3] J. R. Binkley, and S. Singh, "An Algorithm for Anomaly-based Botnet Detection", *Proceedings of USENIX Steps to Reducing Unwanted Traffic on the Internet Workshop (SRUTI)*, July 2006, pp. 43–48
[4] E. Cooke, F. Jahanian and D. McPherson, "The Zombie Roundup: Understanding, Detecting, and Disrupting Botnets", *In Proceedings of Usenix Workshop on Stepts to Reducing Unwanted Traffic on the Internet (SRUTI 05)*. Cambridge, MA, July 2005, pp. 39–44.
[5] F. C. Freiling, T. Holz and G. Wicherski, "Botnet Tracking: Exploring a Root-Cause Methodology to Prevent Distributed Denial-of-Service Attacks", Technical Report AIB-2005-07, RWTH Aachen University, April 2005.
[6] J. Geobel and T. Holz, "Rishi: Identify Bot Communicated Hosts by IRC Nickname Evaluation",
[7] J. Greensmith and U. Aickelin, "Dendritic Cells for SYN Scan Detection", *Proceedings of the Genetic and Evolutionary Computation Conference (GECCO 2007)*, pp. 49–56
[8] J. Greensmith, U. Aickelin and G. Tedesco, "Information Fusion for Anomaly Detection with the Dendritic Cell Algorithm", Accepted for the Special Issue on Biologically Inspired Information Fusion; To be appear in International Journal of Information Fusion, Elservier, 2007.
[9] J. Greensmith, "The Dendritic Cell Algorithm", PhD Thesis, University of Nottingham, 2007.
[10] The Honeynet Project, "Know your enemy: Tracking botnets", http://www.honeynet.org/papers/bots/, March 2005.
[11] N. Ianelli and A. Hackworth, "Botnets as a Vehicle for Online Crime. CERT Coordination Center", 2005.
[12] IceChat - IRC Client: http://www.icechat.net/, Accessed on $10^{t}h$ March 2008.
[13] C. Kalt, Internet Relay Chat: Architecture. Request for Comments: RFC 2810, April 2000.
[14] M. Lutz and G. Schuler, " Immature, semi-mature and fully mature dendritic cells: which signals induce tolerance or immunity? ", Trends in Immunology, 23(9):9911045, 2002.
[15] S. Racine, "Analysis of Internet Relay Chat Usage by DDoS Zombies", Master's Thesis. Swiss Federal Institute of Technology Zurich, April 2004.
[16] J. Twycross and U. Aickelin, Biological Inspiration for Artificial Immune Systems, Proc. of the 6th International Conference on Artificial Immune Systems, Santos/SP, Brazil, August 2007.